\documentclass[10pt,twocolumn,letterpaper]{article}

\usepackage{cvpr}
\usepackage{times}
\usepackage{epsfig}
\usepackage{graphicx}
\usepackage{amsmath}
\usepackage{amssymb}
\usepackage{multirow}
\usepackage{booktabs}
\usepackage{color}
\usepackage{authblk}
\usepackage[utf8]{inputenc}
\usepackage[T1]{fontenc}

\usepackage[breaklinks=true,bookmarks=false]{hyperref}

\cvprfinalcopy 

\def\cvprPaperID{1943} 

\begin{document}
	
	\title{A Skeleton-bridged Deep Learning Approach for Generating Meshes \\ of Complex Topologies from Single RGB Images}
	
	\author[1]{Jiapeng Tang $^{*}$}
	\author[2]{Xiaoguang Han \thanks{Equal contributions}}
	\author[1]{Junyi Pan}
	\author[1]{Kui Jia \thanks{Corresponding author}}
	\author[3]{Xin Tong}
	\affil[1]{School of Electronic and Information Engineering, South China University of Technology}
	\affil[2]{Shenzhen Research Institute of Big Data, the Chinese University of Hong Kong (Shenzhen)}
	\affil[3]{Microsoft Research Asia}
	\affil[ ]{{\tt\small msjptang@mail.scut.edu.cn, \tt\small hanxiaoguang@cuhk.edu.cn, eejypan@mail.scut.edu.cn, \tt\small kuijia@scut.edu.cn, \tt\small  xtong@microsoft.com}}
	
	\maketitle
	\thispagestyle{empty}
	\pagestyle{empty}
	
	\begin{abstract}
		This paper focuses on the challenging task of learning 3D object surface reconstructions from single RGB images. Existing methods achieve varying degrees of success by using different geometric representations. However, they all have their own drawbacks, and cannot well reconstruct those surfaces of complex topologies. To this end, we propose in this paper a skeleton-bridged, stage-wise learning approach to address the challenge. Our use of skeleton is due to its nice property of topology preservation, while being of lower complexity to learn. To learn skeleton from an input image, we design a deep architecture whose decoder is based on a novel design of parallel streams respectively for synthesis of curve- and surface-like skeleton points. We use different shape representations of point cloud, volume, and mesh in our stage-wise learning, in order to take their respective advantages. We also propose multi-stage use of the input image to correct prediction errors that are possibly accumulated in each stage. We conduct intensive experiments to investigate the efficacy of our proposed approach. Qualitative and quantitative results on representative object categories of both simple and complex topologies demonstrate the superiority of our approach over existing ones. We will make our ShapeNet-Skeleton dataset publicly available.
	\end{abstract}
	
	\section{Introduction}
	\label{SecIntro}
	
	\begin{figure}[t]
		\centering
		\includegraphics[width=0.8\linewidth]{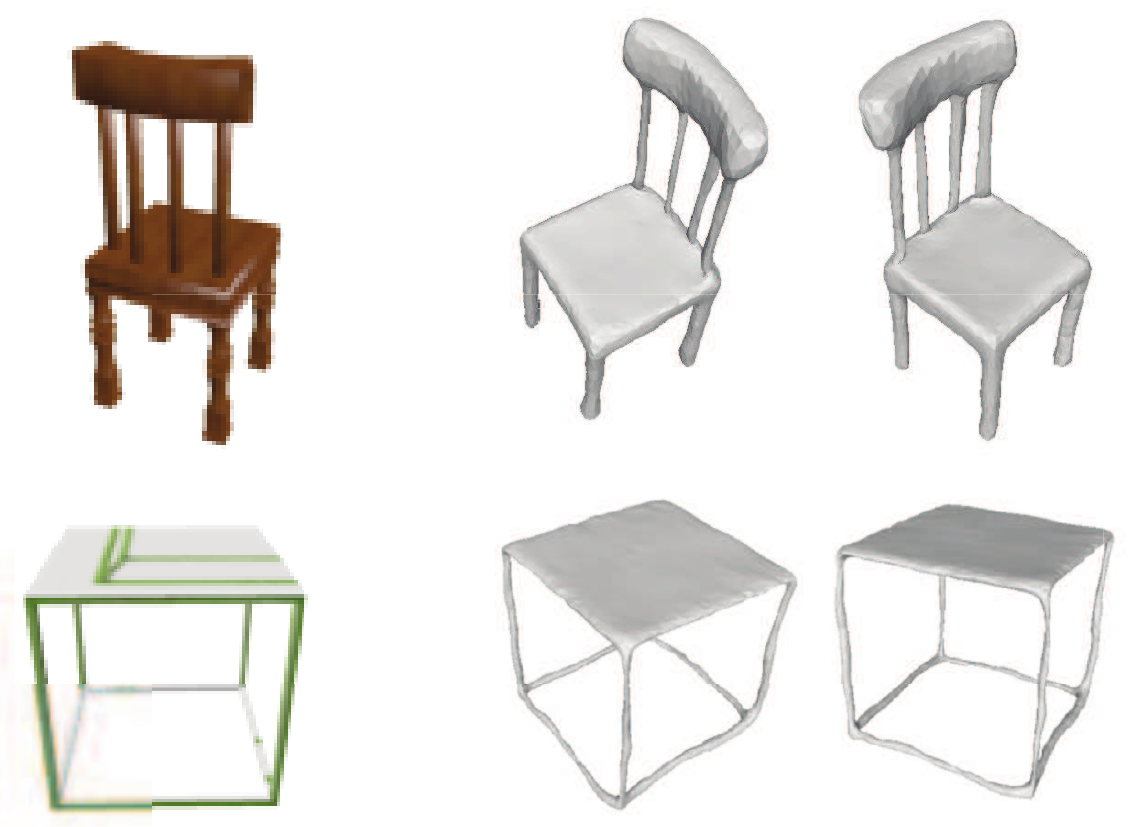}
		\caption{Our proposed approach can generate a closed surface mesh from a single view RGB image, by correctly recovering the complex topology.}
		\label{fig:show}
	\end{figure}
	
	Learning 3D surface reconstructions of objects from single RGB images is an important topic from both the academic and practical perspectives.
	It plays fundamental roles in applications such as augmented reality and image editing. It also connects with the more traditional research of 3D vision ~\cite{hartley2000multiple, fuentes-pacheco2015visual, häming2010structure}. This inverse problem is extremely challenging due to the arbitrary shapes of different object instances and their possibly complex topologies. Recent methods \cite{choy20163d, girdhar2016learning, tatarchenko2017octree,  wu2016learning, groueix2018atlasnet, kato2018neural, sinha2017surfnet, fan2017point, lin2017learning,wang2018pixel2mesh, pan2018residual} leverage the powerful learning capacities of deep networks, and achieve varying degrees of success by using different shape representations, e.g., volume, point cloud, or mesh. These methods have their own merits but also have their respective drawbacks. For example, volume-based methods \cite{choy20163d, girdhar2016learning, wu2016learning} exploit the establishment of Convolutional Neural Networks (CNNs) ~\cite{simonyan2015very, krizhevsky2012imagenet,szegedy2015going,he2016deep}, and simply extend CNNs its 3D versions to generate volume representations of 3D shapes; however, both of their computational and memory complexities are high enough which prohibit them to be deployed to generate high-resolution outputs. On the other hand, point cloud based methods \cite{fan2017point, lin2017learning} are by nature difficult to generate smooth and clean surfaces.
	
	Given the fact that mesh representation is a more efficient, discrete approximation of the continuous manifold of an object surface, a few recent methods \cite{fan2017point, lin2017learning} attempt to directly learn mesh reconstructions from single input images. These methods are inherently of mesh deformation, since they assume that an initial meshing over point cloud is available; for example, they typically assume unit square/sphere as the initial mesh. In spite of the success achieved by these recent methods, they still suffer from generating surface meshes of complex topologies, e.g., those with thin structures as shown in Fig \ref{fig:show}.
	
	To this end, we propose in this paper a \emph{skeleton-bridged, stage-wise} deep learning approach for generating mesh reconstructions of object surfaces from single RGB images. We particularly focus on those object surfaces with complex topologies, e.g., chairs or tables that have local, long and thin structures. Our choice of the \emph{meso-skeleton} \footnote{Skeletal shape representation is a kind of medial axis transform (MAT). While the MAT of a 2D shape is a 1D skeleton, for a 3D model, the MAT is generally composed of 2D surface sheets. The skeleton composed of skeletal curves and skeletal sheets (i.e., medial axes) is generally called meso-skeleton.} is due to its nice property of topology preservation, while being of lower complexity to learn when compared with learning the surface meshes directly. Our proposed approach is composed of three stages. The first stage learns to generate skeleton points from the input image, for which we design a deep architecture whose decoder is based on a novel, parallel design of CurSkeNet and SurSkeNet, which are respectively responsible for the synthesis of curve- and surface-like skeleton points. To train CurSkeNet and SurSkeNet, we compute skeletal shape representations for instances of ShapeNet \cite{chang2015shapenet}. \emph{We will make our ShapeNet-Skeleton dataset publicly available.} In the second stage, we produce a base mesh by firstly converting the obtained skeleton to its coarse volumetric representation, and then refining the coarse volume using a learned 3D CNN, where we adopt a strategy of independent sub-volume synthesis with regularization of global structure, in order to reduce the complexity of producing high-resolution volumes. In the last stage, we generate our final mesh result by extracting a base mesh from the obtained volume \cite{lorensen1987marching}, and deforming vertices of the base mesh using a learned Graph CNN (GCNN) \cite{kipf2017semi,defferrard2016convolutional,boscaini2016learning,scarselli2009graph}. Learning and inference in three stages of our approach are based on different shape representations, which take the respective advantages of point cloud, volume, and mesh. We also propose multi-stage use of the input image to correct prediction errors that are possibly accumulated in each stage. We conduct intensive ablation studies which show the efficacy of stage-wise designs of our proposed approach.
	
	We summarize our main contributions as follows.
	\begin{itemize}
		\item Our approach is based on an integrated stage-wise learning, where learning and inference in different stages are based on different shape representations by taking the respective advantages of point cloud, volume, and mesh. We also propose multi-stage use of the input image to correct prediction errors that are possibly accumulated in each stage.
		
		\item We propose in this paper a skeleton-bridged approach for learning object surface meshes of complex topologies from single RGB images. Our use of skeleton is due to its nice property of topology preservation, while being of lower complexity to learn. We design a deep architecture for skeleton learning, whose decoder is based on a novel design of parallel streams respectively for the synthesis of curve- and surface-like skeleton points. To train the network, we prepare ShapeNet-Skeleton dataset and will make it publicly available.
		
		\item We conduct intensive ablation studies to investigate the efficacy of our proposed approach. Qualitative and quantitative results on representative object categories of both simple and complex topologies demonstrate the superiority of our approach over existing ones, especially for those objects with local thin structures.
	\end{itemize}
	
	\begin{figure*}[t]
		\begin{center}
			\includegraphics[scale=0.55]{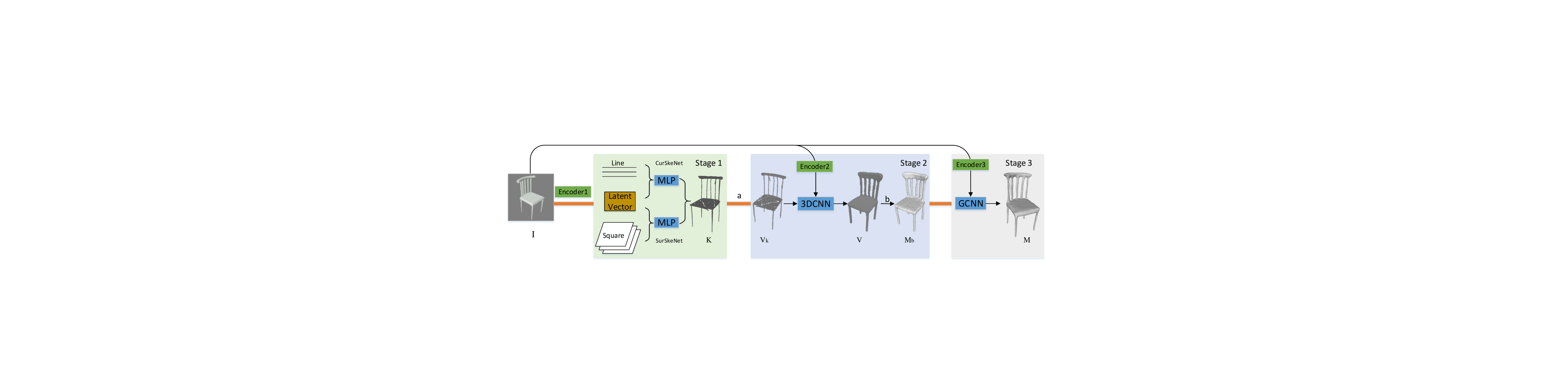}
			\caption{Our overall pipeline. Given an input image $I$, we employ two parallel MLPs to infer skeletal points $K$ in stage one. After converting $K$ to a coarse volume $V_k$,  we refine $V_k$ to get $V$ by 3D CNN and extract base mesh $M_b$ from $V$ in stage two. We further optimize vertices of $M_b$ using GCNN to acquire a final mesh $M$. The operation $a$ means voxelization and the operation $b$ stands for Marching Cubes. }
			\label{fig:overall}
		\end{center}
	\end{figure*}
	
	\section{Related Works}
	In this section, we only focus on the related works about deepnets-based algorithms for fully object reconstruction.
	The literature reviews are studied in the following three aspects.
	
	\noindent \textbf{Volume-based Generator} Voxels, extended from pixels, are usually used in the form of binary values or signed distances to represent a 3D shape. Because of its regularity, most of existing deepnets-based shape analysis ~\cite{wu20153d, brock2016generative} or shape generation ~\cite{choy20163d, girdhar2016learning, wu2016learning} methods adopt it as the primary representation. For example, the work of ~\cite{choy20163d} combines 3D convolutions with long short-term memory (LSTM) units to achieve volumetric grid reconstruction from single-view or multi-view RGB images.
	A 3D auto-encoder is trained in ~\cite{girdhar2016learning}, whose decoder part is used to construct the mapping from a 2D image to a 3D occupancy grid.
	These methods tend to predict a low-resolution volumetric grid due to the high computational cost of 3D convolution operators. Based on the observation that only a small portion of regions around the boundary surface contain the shape information, the Octree representation has been adopted in recent shape analysis works ~\cite{riegler2017octnet, wang2017cnn}.   A convolutional Octree decoder is also designed in ~\cite{tatarchenko2017octree} to support high-resolution reconstruction with a limited time and memory cost. In our work, we aim to generate the surface mesh of the object instead of its solid volume. As its efficiency and topology-insensitivity, we also leverage volumetric-based generator to convert the inferred skeletal point cloud to a solid volume, effectively bridging the gap between the skeleton and the surface mesh.
	
	\noindent \textbf{Surface-based Generator} Point cloud, sampled from the object' surface and formed by a set of points,
	is one of the most popular representations of 3D shapes. Fan et al. ~\cite{fan2017point} proposes the first point could generation neural network, which is built upon a deep regression model trained with the loss functions that evaluate the similarity of two unordered point set, such as chamfer distance. Although the rough shape can be captured, the generated points are placed sparse and scattered.
	Multi-view depth maps are used as the output representation in ~\cite{lin2017learning}, which are generated with image generative models and then fused to give rise to a dense point cloud. Nevertheless, the predicted points are still of low accuracy.
	Mesh, as the most natural discretization of a manifold surface, has been widely used in many graphics applications. Due to its irregular structure, CNN is difficult to be directly applied to mesh generation. To alleviate this challenge, the methods of ~\cite{kato2018neural, wang2018pixel2mesh} take an extra template mesh as input and attempt to learn the deformations to approximate the target surfaces. Limited to the requirement of an initial mesh, they cannot deal with topology-free reconstruction. Another recent method, called Atlasnet ~\cite{groueix2018atlasnet}, proposes to deform multiple 2D planar patches to cover the surface of the object. Residual prediction and progressive deformation are adopted in  ~\cite{pan2018residual}, which decrease the complexity of learning and make more details added. It is free of complex topology yet causes severe patch overlaps and holes. In our work, we aim not only to generate a clean mesh but also to capture the correct topology. To do so, we firstly borrow the idea in ~\cite{groueix2018atlasnet} to infer the meso-skeleton points, which are then converted to a base mesh. Finally, the method of ~\cite{wang2018pixel2mesh} is further adopted for generating geometric details.
	
	\noindent \textbf{Structure Inference} Instead of estimating geometric shapes, many recent works attempt to recover the 3D structures of objects. 
	From a single image, Zou et al. ~\cite{Zou_2017_ICCV} presents a primitive recurrent neural network to sequentially predict a set of cuboid primitives to approximate the shape structure. A recursive decoder is proposed in \cite{li_sig17} to generate shape parts and infer reasonable high-level structure information including part connectivity, adjacency and symmetry relation. This is further exploited in ~\cite{niu2018im2struct} for image-based structure inference. However, the cuboids are hard to be used for fitting curved shapes. In addition, these methods also require a large human-labeled dataset. We use meso-skeleton, a point cloud, to represent the shape structure which is easier to be obtained from the ground truth. The usage of parametric line and square elements also eases the approximation of the diverse local structures.
	
	\section{The Proposed Approach}
	\label{SecFramework}
	
	We first overview our proposed skeleton-bridged approach for generating a surface mesh from an input RGB image, before explaining the details of stage-wise learning. Given an input image $I$ of an object, our goal is to recover a surface mesh $M$ that ideally captures the \emph{possibly complex} topology of 3D shape of the object. This is an extremely challenging inverse task; existing methods \cite{kato2018neural,wang2018pixel2mesh,groueix2018atlasnet} may only achieve partial success for objects with relatively simple topologies. To address the challenge, our key idea in this work is to bridge the mesh generation of object surface via learning of meso-skeleton. As discussed in Section \ref{SecIntro}, the rationale is that skeletal shape representation preserves the main topological structure of a shape, while being of lower complexity to learn.
	
	More specifically, our mesh generation process is composed of the following three stages. In the first stage, we learn an encoder-decoder architecture that maps $I$ to its meso-skeleton $K$, represented as a compact point cloud. In the second stage, we produce a volume $V$ from $K$ by firstly converting $K$ to its coarse volumetric representation $V_k$, and then refining $V_k$ using a learned 3D CNN (e.g., of the style \cite{han2017high}). In the last stage, we generate the final output mesh $M$ by extracting a base mesh $M_b$ from $V$, and further optimizing vertices of $M_b$ using a learned graph CNN \cite{scarselli2009graph}. Each stage owns its own image encoder, and thus inferences in all the three stages are guided by the input image $I$. Fig ~\ref{fig:overall} illustrates the whole pipeline of our approach.
	
	\subsection{Learning of Meso-Skeleton}
	\label{SecSkeletonLearning}
	
	As defined in Section \ref{SecIntro}, the meso-skeleton of a shape is represented as its medial axis, and the medial axis of a 3D model is made up of curve skeletons and median sheets, which are adaptively generated from local regions of the shape. In this work, we utilize the skeleton representation introduced in ~\cite{wu2015deep}, i.e., a compact point cloud. Fig ~\ref{fig:com_ske} shows an example of skeleton that we aim to recover.
	
	\noindent\textbf{The ShapeNet-Skeleton dataset} Training skeletons are necessary in order to learn to generate a skeleton from an input image. In this work, we prepare training data of skeleton for ShapeNet \cite{chang2015shapenet} as follows: 1) for each 3D polygonal model in ShapeNet, we convert it into a point cloud; 2) we extract meso-skeleton points using the method of ~\cite{wu2015deep}; 3) we classify each skeleton point as either curve-like or surface-like categories, based on principle component analysis of its neighbor points. \emph{We will make our ShapeNet-Skeleton dataset publicly available. }
	
	
	\noindent\textbf{CurSkeNet and SurSkeNet} Given the training skeleton points for the object in each image, we design an encoder-decoder architecture for skeleton learning, where the input $I$ is firstly encoded to a latent vector that is then decoded to a point cloud of skeleton. Our encoder is similar to those in existing methods of point set generation, such as ~\cite{fan2017point, groueix2018atlasnet}. In this work, we use ResNet-18 ~\cite{he2016deep} as our image encoder. Our key contribution is a novel design of decoder architecture that will be presented shortly. We note that one may think of using existing methods ~\cite{fan2017point, groueix2018atlasnet} to generate $K$ from $I$; however, they tend to fail due to the complex, especially thin, structures of skeletons, as shown in Fig ~\ref{fig:com_ske}.  Our decoder is based on two parallel streams of CurSkeNet and SurSkeNet, which are designed to synthesize the points at curve-shaped and surface-shaped regions respectively.  Both CurSkeNet and SurSkeNet are based on multilayer perceptrons (MLPs)  with the same settings as in AtlasNet ~\cite{groueix2018atlasnet}, including 4 fully-connected layers with the respective sizes of 1024, 512, 256, and 3, where the non-linear activation functions are ReLU for the first 3 layers and tanh for the last layer. Our SurSkeNet learns to deform a set of 2D primitives defined on the open unit square $[0,1]^2$, producing a local approximation of the desired sheet skeleton points. Our CurSkeNet learns to deform a set of 1D primitives defined on the open unit line $[0,1]$; it thus conducts affine transformations on them to form curves, and learns to assemble generated curves to approximate the curve-like skeleton part. In our current implementation, we use 20 line primitives in CurSkeNet and 20 square primitives in SurSkeNet. In Section ~\ref{SecCompSke}, we conduct ablation studies that verify the efficacy of our design of CurSkeNet and SurSkeNet.

	\noindent\textbf{Network Training} We use training data of curve-like and surface-like skeleton points to train CurSkeNet and SurSkeNet. The learning task is essentially of point set generation. Similar to ~\cite{groueix2018atlasnet}, we use the Chamfer Distance (CD) as one of our loss functions. The CD loss is defined as:
	\begin{align}
	{\cal{L}}_{cd} = \sum_{x\in K} \min_{y\in K^{*}}\|x-y\|_2^2 + \sum_{y\in K^{*}}\min_{x\in K}\|x-y\|_2^2 ,
	\end{align}
	where $\{x \in K\}$ and $\{y \in K^{*}\}$ are respectively the sets of predicted and training points. Besides, to ensure local consistency, regularizer of Laplacian smoothness is also used for generation of both curve- and surface-like points. It is defined as:
	\begin{align}
	{\cal{L}}_{lap} = \sum_{x \in K}\Big\|x -\frac{1}{|\mathcal{N}(x)|}\sum_{p\in \mathcal{N}(x)} p\Big\|_{2} ,
	\end{align}
	where $\mathcal{N}(x)$ is the neighbor of point $x$.
	

	\subsection{From Skeleton to Base Mesh}
	\label{SecBaseMesh}
	
	We present in this section how to generate a base mesh $M_b$ from the obtained skeleton $K$. To do so, a straightforward approach is to coarsen $K$ to a volume directly with hand-crafted methods, and then to produce the base mesh using the method of Marching Cubes ~\cite{lorensen1987marching}. However, such an approach may accumulate stage-wise prediction errors. Instead, we rely on the original input $I$ to correct the possible stage-wise errors, by firstly converting $K$ to its volumetric representation $V_k$, and then using a trained 3D CNN for a finer and more accurate volumetric shape synthesis, resulting in a volume $V$. Base mesh $M_b$ can then be obtained by applying Marching Cubes to the finer $V$.
	
	\begin{figure}[t]
		\begin{center}
			\includegraphics[scale=0.28]{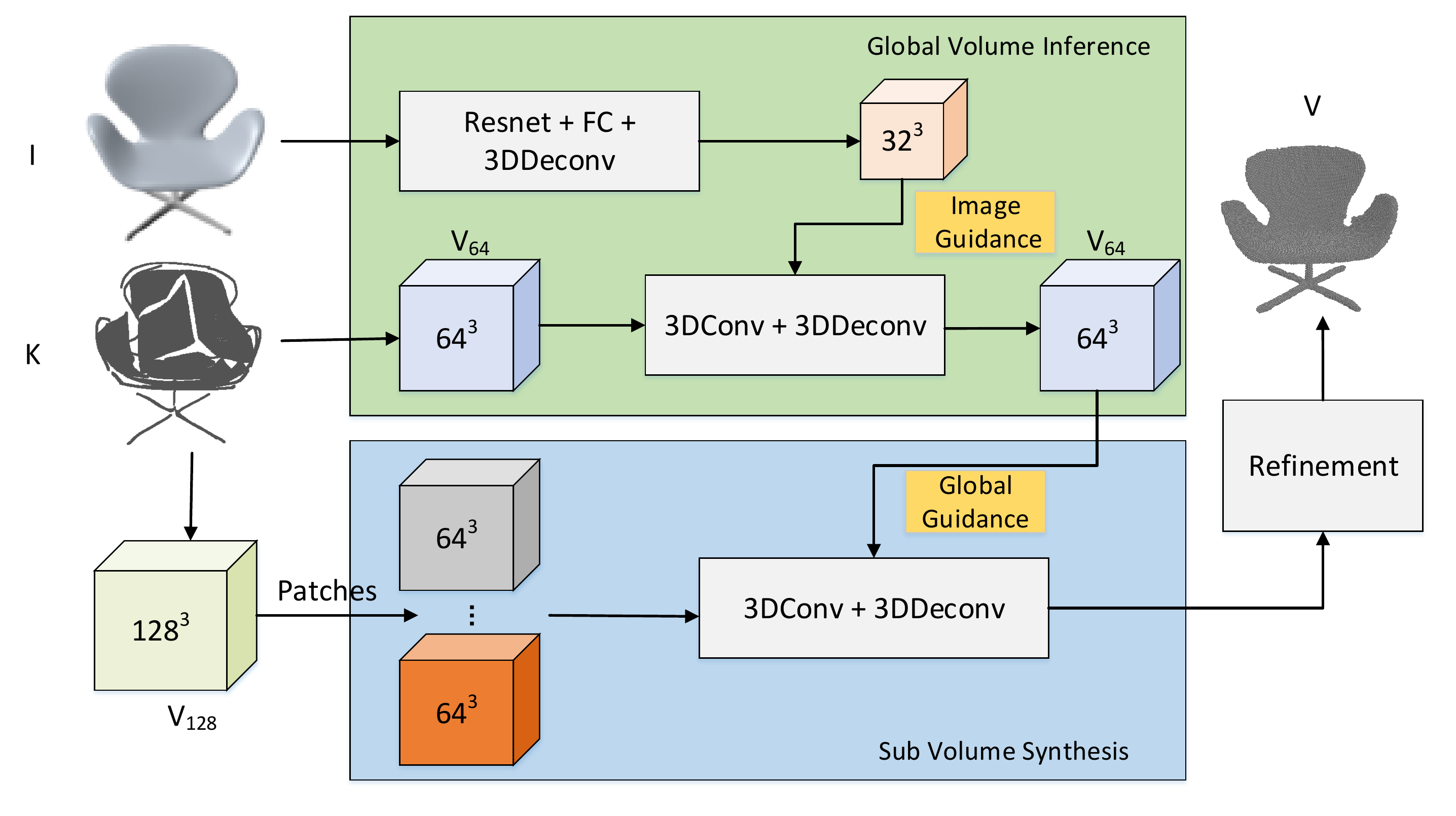}
			\caption{The pipeline of our high-resolution skeletal volume synthesis method. We convert the inferred skeletal points $K$ to low-resolution volume $V_{64}$ and high-resolution volume $V_{128}$ in parallel. Given $V_{64}$, $V_{128}$ paired with the input image $I$, a global-guided sub-volume synthesis network is proposed to output a refined volume of  $V_{128}$. It consists of two subnetworks: one network generates a coarse skeletal volume from $I$ and $V_{64}$ while the other enhances  $V_{128}$ locally patch by patch under the guidance of the output from the first network.}
			\label{fig:local_global_alg}
		\end{center}
	\end{figure}
	
\begin{figure}[h]
		\begin{center}
			\includegraphics[scale=0.55]{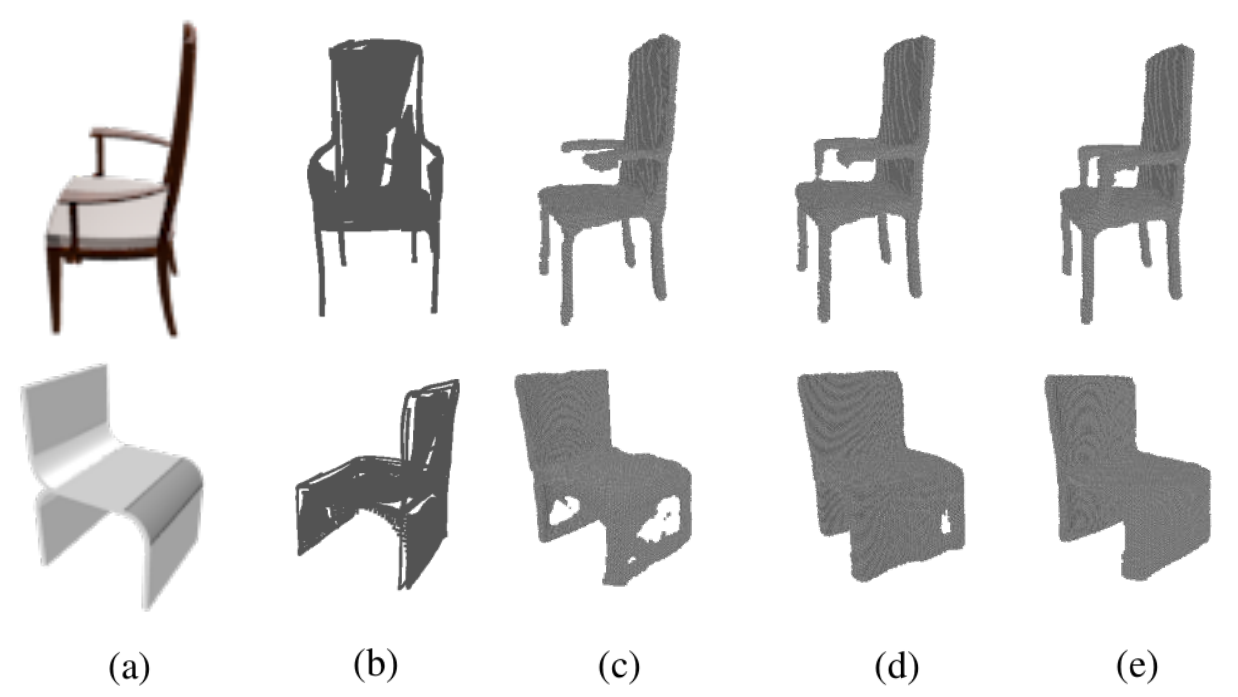}\\
			\caption{(a)Input images; (b)Inferred skeletal points; (c)sub-volume synthesis only;
				(d) adding global guidance; (e) adding image guidance.}
			\label{fig:local_global_comp}
		\end{center}
	\end{figure}

	\noindent\textbf{Sub-volume Synthesis with Global Guidance} To preserve the topology captured by $K$, a high-resolution volume representation is required. However, this is not easy to satisfy due to the expensive computational cost of 3D convolution operations. OctNet ~\cite{riegler2017octnet} may alleviate the computational burden, it is however complex and difficult to implement. We instead partition the volume space into overlapped sub-volumes, and conduct refinement on them in parallel. We also follow ~\cite{han2017high} and employ a global guidance to preserve spatial consistency across sub-volumes. More specifically, we firstly convert $K$ to two volumes of varying scales, denoted as $V^{l}_k$ and $V^{h}_k$. We set  $| V^{l}_k |  = 64^3$ and $| V^{h}_k |  = 128^3$ in this work. We use two networks of 3D CNNs for global and local synthesis of skeletal volumes. The global network is trained to refine $V^{l}_k$ and generate a skeletal volume $V^{l}$ of the size $64^3$. The local network takes as inputs sub-volumes of the size $64^3$, which are uniformly cropped from $V^{h}_k$, and then conduct their refinement individually. Both of our global and local refinement networks are based on 3D U-Net architecture \cite{ronneberger2015u}. When refining each sub-volume of $V^{h}_k$, the corresponding $32^3$-sized sub-volume of $V^{l}$ is concatenated to provide structural regularization. The overall pipeline of our method is shown in Fig ~\ref{fig:local_global_alg}. As seen in Fig ~\ref{fig:local_global_comp}, our method not only supports high-resolution synthesis but also preserves global structure.
	
	\noindent\textbf{Image-guided Volume Correction} To correct the possiblely accumulated prediction errors from the stage of skeleton generation, we reuse the original input $I$ by learning an independent encoder-decoder network, which is trained to map $I$ to a $32^3$-sized volume. We use ResNet-18 as the encoder and several 3D de-convolution layers as the decoder. The output of the decoder is incorporated into the aforementioned global synthesis network, aiming for a more accurate $V^{l}$, which ultimately contributes to the generation of a better $V$. From the perspective of learning task for generating 3D volumes from single images \cite{choy20163d,girdhar2016learning,wu2016learning,tatarchenko2017octree}, our method is superior to existing ones by augmenting with an additional path of skeleton inference. As shown in Fig ~\ref{fig:local_global_comp}, our usage of $I$ for error correction greatly improves the synthesis results.
	
	\noindent\textbf{Base Mesh extraction} Given $V$, we use Marching Cubes ~\cite{lorensen1987marching} to produce the base mesh $M_b$, which ideally preserves the same topology as that of the skeleton $K$. Because $V$ is in high resolution, $M_b$ would contain a large number of vertices and faces. To reduce the computational burden of the last stage, we apply QEM algorithm \cite{kobbelt1998general} on $M_b$ to get a simplified mesh for subsequent processing.
	
	\subsection{Mesh Refinement}
	
	\begin{figure}[h]
		\begin{center}
			\includegraphics[width=0.90\linewidth]{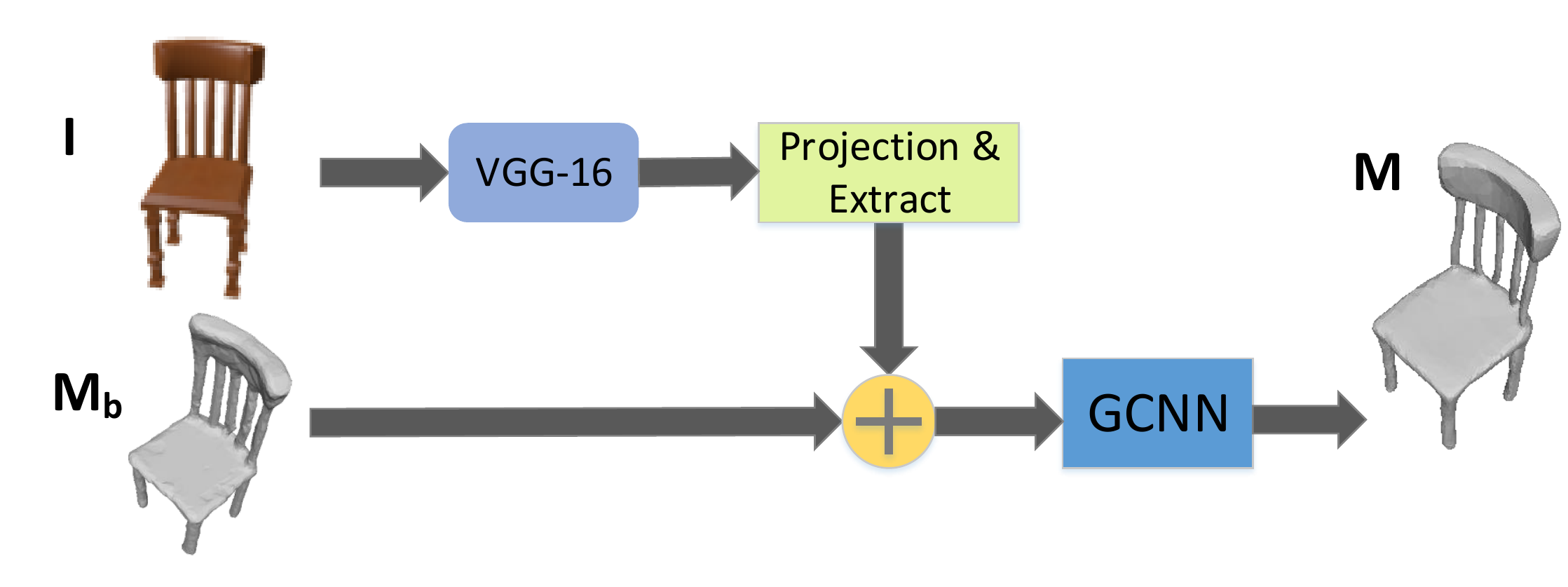}
			\caption{Our mesh refinement network. Given an image $I$ and an initial mesh $M_b$, we concatenate pixel-wise features of $I$ (extracted by VGG-16) to vertices' coordinates and form vertex-wise features which are followed by a graph-CNN to generate the geometric details. }
			\label{fig:mesh_refine_alg}
		\end{center}
	\end{figure}
	
	We have up to now the base mesh $M_b$ that captures the topology of the underlying object surface, but may lack surface details. To compensate $M_b$ with surface details, we take the approach of mesh deformation using graph CNNs  \cite{kipf2017semi,defferrard2016convolutional,boscaini2016learning,scarselli2009graph}.
	
	\begin{figure*}[h]
		\begin{center}
			\includegraphics[width=0.93\linewidth]{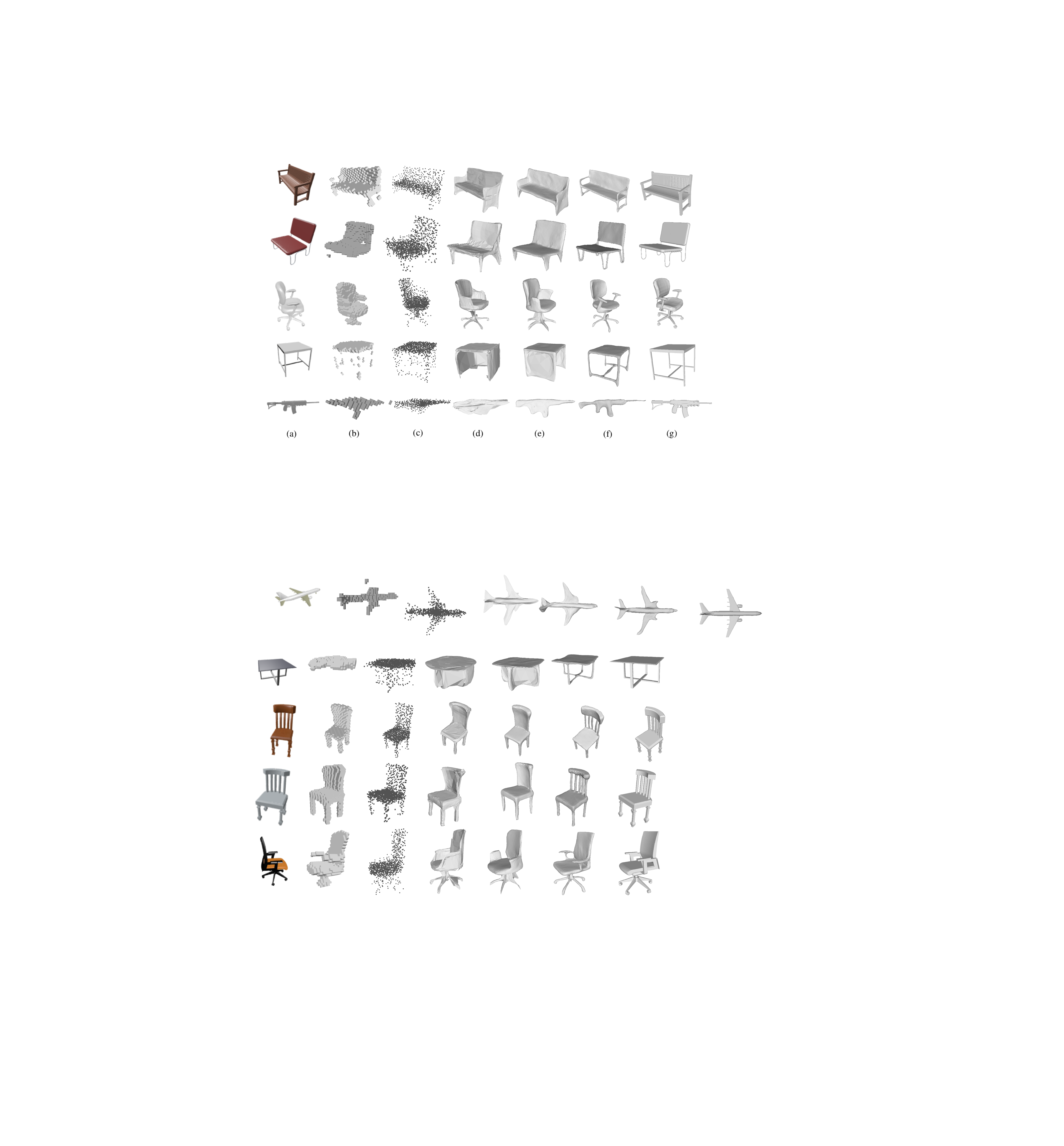}\\
		\end{center}
		\caption{ (a)Input images; (b)R2N2;  (c)PSG;  (d)AtlasNet; (e)Pixel2Mesh;  (f)Ours;  (g)Ground truth }
		\label{fig:comp_recon}
	\end{figure*}
	\noindent\textbf{Mesh Deformation using Graph CNNs} Take $M_b$ as the input, our graph CNN is simply composed of a few graph convolutional layers, each of which apply spatial filtering operation to local neighborhood associated with each vertex point of  $M_b$.  The graph-based covolutional layer is defined as:
	\begin{align}
	h_{p}^{l+1} = w_{0}h_{p}^{l} + \sum_{q\in\mathcal{N}(p)}w_{1}h_{q}^{l} ,
	\end{align}
	where $h_{p}^{l}$, $h_{p}^{l+1}$ are the feature vectors on the vertex $p$ before and after applying a convolution operation, and $\mathcal{N}(p)$ is the neighbor of $p$. $w_{0}$ and $w_{1}$ are the learnable parameter matrices that are applied to all vertices.
	
	Similar to ~\cite{wang2018pixel2mesh}, we also concatenate pixel-wise VGG features extracted from $I$ with coordinates of the corresponding vertices to enhance learning. We again use CD loss to train our graph CNN. Several smoothness terms are also added to regularize the mesh deformation. One is edge regularization, used to avoid large deformations, by restricting the length of output edges. Another one is normal loss, used to guarantee the smoothness of the output surface. The geometric details commonly exist at the regions where the normals are changed obviously. Regarding this fact, to guide the GCNN to better learn the surface in those areas, we accordingly construct weighted loss functions. Fig ~\ref{fig:mesh_refine_alg} shows the efficacy of this weighting strategy, where the sharp edges are better synthesized.

	\section{Experiments}
	
	\noindent \textbf{Dataset} To support the training and testing of our proposed approach, we collect 17705 3D shapes from five categories in ShapeNet ~\cite{chang2015shapenet}: plane(1000), bench(1816), chair(5380), table(8509), firearm(1000). The dataset is split into two parts, $80\%$ shapes are used for training and the other for testing. We take as the inputs the rendered images provided by ~\cite{choy20163d}, where each 3D model is rendered into 24 RGB images. Each shape in the dataset is converted to a point cloud ($10,000$ points are sampled on the surface) as the ground truth for mesh refinement network.
	
	\noindent \textbf{Implementation details} The input images are all in the size of 224*224. We train CurSkeNet and SurSkeNet using a batch size of 32 with a learning rate of 1e-3 (dropped to 3e-4 after 80 epochs) for 120 epochs. The skeletal volume refinement network is trained in three steps: 1) the global volume inference network is trained alone with learning rate 1e-4 for 50 epochs(dropped to 1e-5 after 35 epochs); 2) we train the sub-volume synthesis network with learning rate 1e-5 for 10 epochs; 3) the entire network is fine-tuned. The mesh refinement network is trained with learning rate 3e-5 for 50 epochs(dropped to 1e-5 after 20 epochs) using a batch size of 1.
	\setlength{\tabcolsep}{4pt}
	\begin{table*}[t]
		\begin{center}
			\begin{tabular}{*{11}{c}}
				\toprule
				\multirow{2}*{Category} & \multicolumn{5}{c}{CD}   & \multicolumn{5}{c}{EMD} \\
				\cmidrule(lr){2-6}\cmidrule(lr){7-11}
				& R2N2 & PSG &  AtlasNet & Pixel2Mesh & Ours 	& R2N2 & PSG &  AtlasNet & Pixel2Mesh & Ours\\
				\midrule
				\midrule
				plane   & 10.434 & 3.824 & 1.529  & 1.890 & \textbf{1.364}   & 11.060  & 13.945  & 8.981 & 7.728  & \textbf{6.026} \\
				bench   & 10.511 & 3.504 & 2.264  & 1.774 & \textbf{1.639}   & 10.555  & 8.053   & 9.143 & 7.083  & \textbf{6.059} \\
				chair   & 4.723  & 2.553 & 1.342  & 1.923 & \textbf{1.002}   & 7.762   & 10.222  & 7.866 & 8.312  & \textbf{5.484} \\
				firearm & 10.176 & \textbf{1.473} & 2.276  & 1.793 & 1.784   & 9.760   & 12.555  & 9.825 & 6.887  & \textbf{6.413} \\
				table   & 12.230 & 5.466 & 1.751 & 2.109  & \textbf{1.321}   & 11.160  & 9.561   & 9.053 & 7.442  & \textbf{5.688} \\
				\midrule
				mean    & 9.615  & 3.364 & 1.832 & 1.898 & \textbf{1.422}   & 10.059  & 10.867  & 8.974 & 7.490 & \textbf{5.934}\\
				\bottomrule
			\end{tabular}
			\caption{Quantitative comparisons of our method against state-of-the-arts. The Chamfer distance($\times$ $10^3$) and Earth Mover's distance($\times$ 0.01) are used. The lower is better on all metrics.}
			\label{Tab:cd_emd}
		\end{center}
	\end{table*}
	\setlength{\tabcolsep}{1.4pt}
	\subsection{Comparisons against State-of-the-Arts}
	\label{SecCompRecon}
	
	We first evaluated our overall pipeline against existing methods on singe-view reconstruction. 3D-R2N2 ~\cite{choy20163d}, PSG~\cite{fan2017point}, AtlasNet~\cite{groueix2018atlasnet}, Pixel2Mesh~\cite{wang2018pixel2mesh} are chosen for their popularity: 3D-R2N2 is one of the most famous volumetric shape generators, PSG is the first point set generator based on a deep regression model, and both AtlasNet and Pixel2Mesh are current state-of-the-art mesh generator. For fair comparison, these  models are retrained under our preprocessed dataset.
	
	\noindent \textbf{Qualitative results} The visual comparisons are shown in Fig \ref{fig:comp_recon}. As seen, 3D-R2N2 always produces low-resolution volumes which cause broken structures. Their results show no surface details either. The point sets regressed by PSG are sparse and scattered, leading to the difficulty of extracting triangular meshes from them. AtlasNet is capable of generating mesh representations without a strong restriction on the shape's topology. Yet, the outputs are of non-closed and suffer from surface self-penetration, which also gives rise to a challenge to convert it to a manifold mesh. Limited to the requirement of a genus-0 template mesh input, Pixel2Mesh is difficult to reach an accurate reconstruction for the objects with complex topologies, as the chairs shown. Our method shows great superiority than the others from the visual appearances, as it generates closed meshes with accurate topologies and more details. For the examples of firearm as shown, our approach also outperforms Pixel2Mesh, which in another aspect, indicates the proposed approach is also good at recovering the shapes with complex structures no mention to topology.
	
	\noindent \textbf{Quantitative results} Similar to Pixel2Mesh ~\cite{wang2018pixel2mesh}, we adopt Chamfer Distance(CD) and Earth Mover's Distance(EMD) to evaluate the reconstruction quality. Both of them are calculated between the point set ($10,000$ points) sampled on the predicted mesh surface and the ground truth point cloud. The quantitative comparison results are reported in Tab \ref{Tab:cd_emd}. Notably, on both metrics, our approach outperforms all the other methods across almost all listed categories, especial on the models with complex topologies like chairs and tables.
	
\begin{figure}[t]
		\vspace{-0.1cm}
		\begin{center}
			\includegraphics[width=0.98\linewidth]{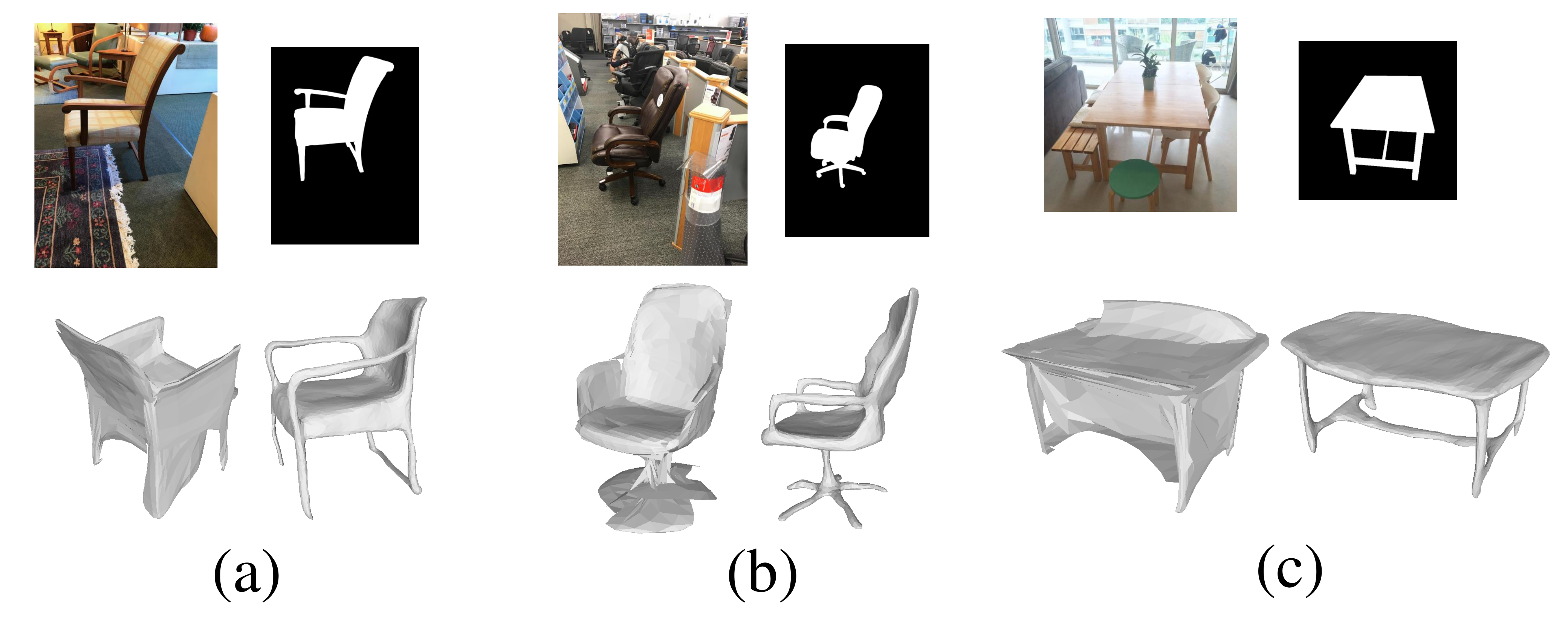}
		\end{center}
		\vspace{-0.5cm}
		\caption{From real photographs and object masks (top row), our method successfully reconstructs 3D object meshes. The results of AtlasNet (left of bottom row) v.s ours (right of bottom row).}
		\vspace{-0.6cm}
		\label{fig:pix3d}
	\end{figure}

	\noindent \textbf{Generalization on real images}  Fig \ref{fig:pix3d} illustrates 3D shapes reconstructed by our method on three real photographs from Pix3D \cite{sun2018pix3d}, where the chairs and tables in the images are manually segmented. The results' quality is similar to the results obtained from synthetic images. As seen in Fig \ref{fig:pix3d} (a), the real-world images has no relation with ShapeNet, while the chair rod can still be well reconstructed. This validates the generalization ability of our method.
	
	\subsection{Ablation Studies on Mesh Generation}
	\label{SecAblaMesh}
	Our whole framework contains multiple stages. In this section, we conduct the ablation studies by alternatively removing one of them, to verify the necessity of each stage.
	
\begin{figure}[h]
		\begin{center}
			\includegraphics[width=0.8\linewidth]{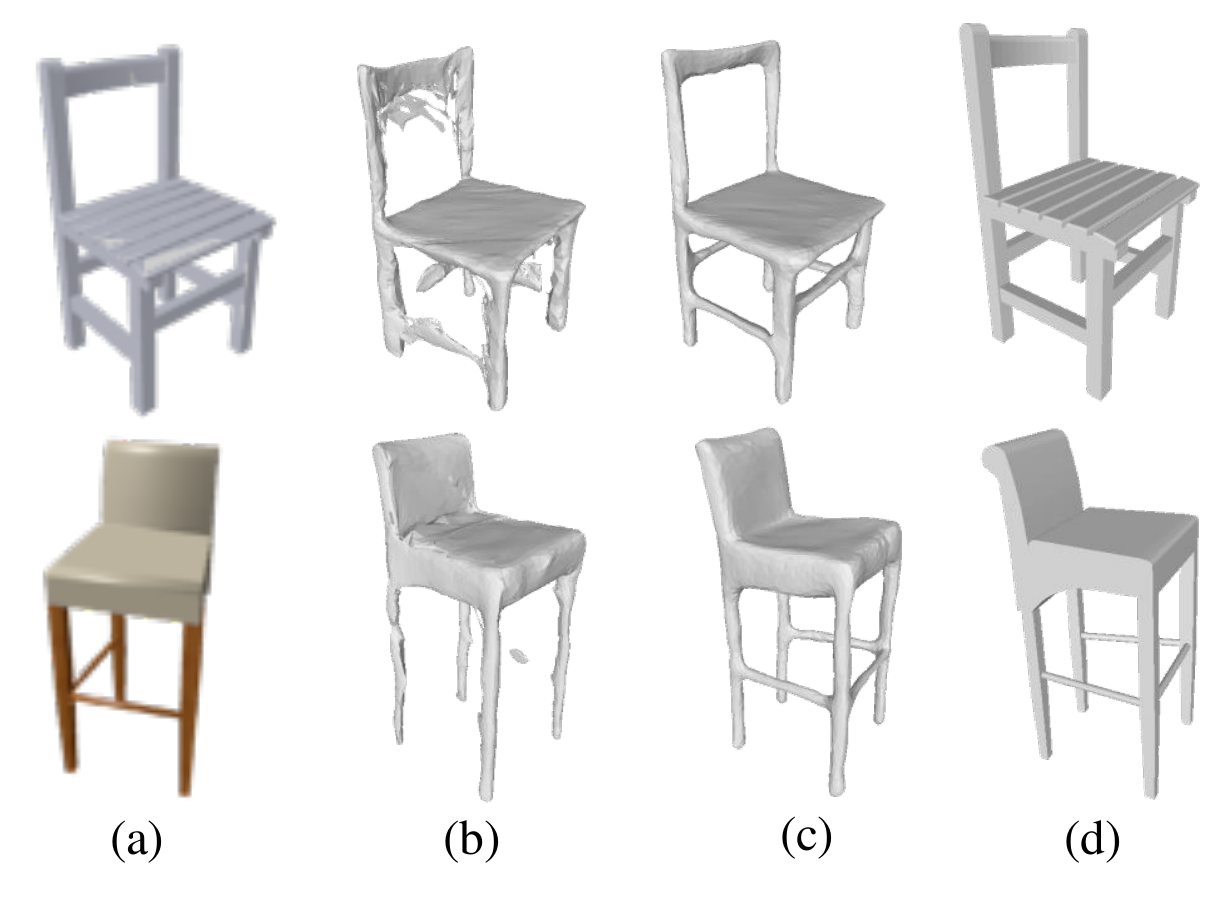}
			\caption{(a)Input images; (b)Final meshes whose base meshes are generated using OGN; (c)The generated meshes of our method; (d)Ground truth.}
			\label{fig:wo_ske}
		\end{center}
	\end{figure}

	\noindent \textbf{w/o skeleton inference} Based on our pipeline, an alternative solution without using skeleton inference is firstly generating a volume directly from the image and then applying our mesh refinement model to output the final result. Then, we implement this approach by using OGN ~\cite{tatarchenko2017octree} as the image-based volume generator, for high-resolution($128^3$) reconstruction.
	This method is compared with ours visually in Fig \ref{fig:wo_ske}.
	As seen, the OGN-based mesh generation method fails to capture the thin structures which causes incorrect topologies. In contrast, our approach gives rise to much better performance. 
	
	\begin{figure*}[t]
		\begin{center}
			\includegraphics[width=0.80\linewidth]{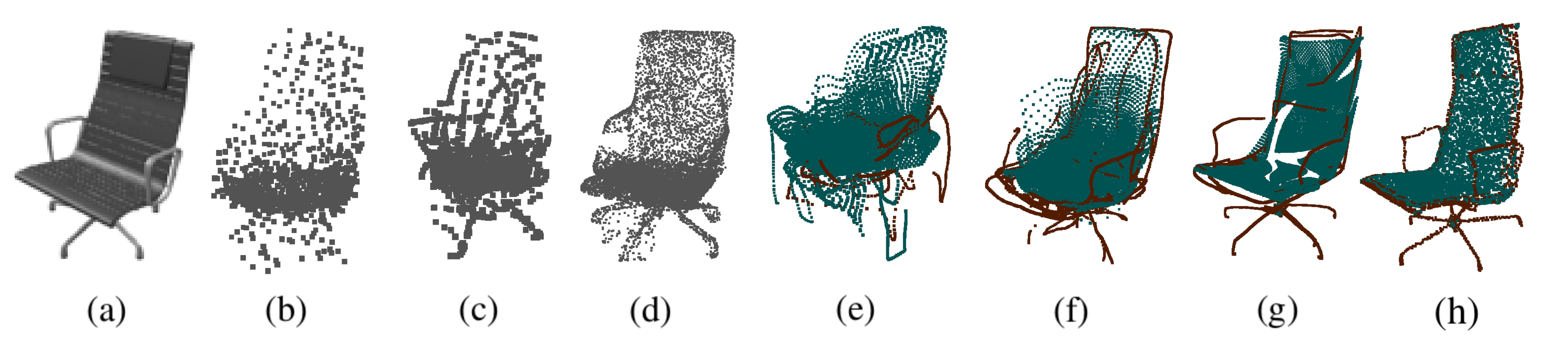}
			\caption{(a)Input images; (b)Point-fitting only; (c)Line-fitting only; (d)Square-fitting only; (e)Line-and-square fitting; (f)Ours w/o laplacian; (g)Ours final; (h)Ground truth.}
			\label{fig:com_ske}
		\end{center}
	\end{figure*}
	
	\begin{figure}[h]
		\begin{center}
			\includegraphics[width=0.85\linewidth]{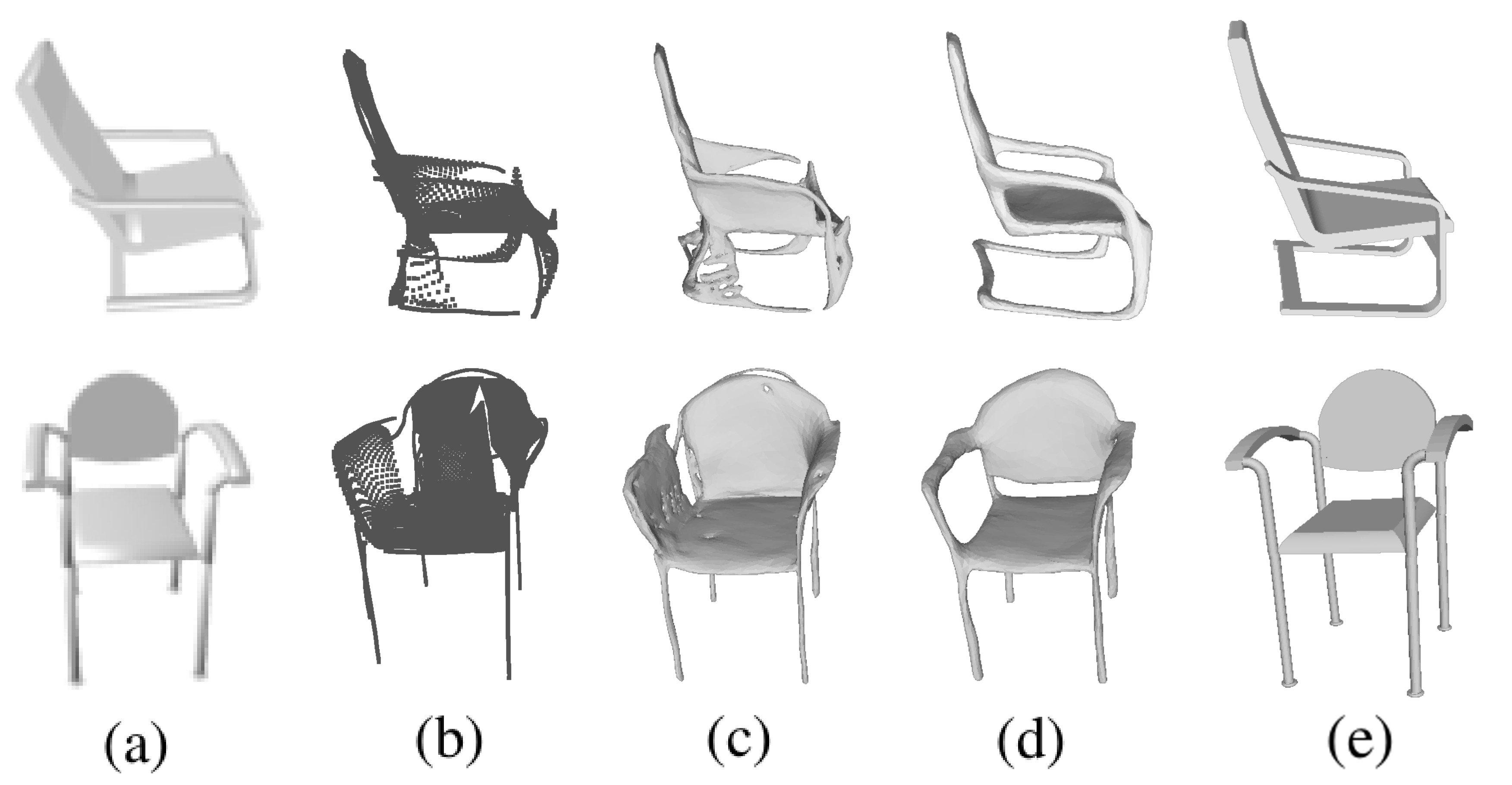}
			\caption{(a)Input images; (b)Inferred skeleton points; (c)The sythesized meshes whose base meshes are extracted from the coarsened skeletal volume using the corrosion techinique; (d) The generated meshes of our method; (e)Ground truth.}
			\label{fig:wo_voxel}
		\end{center}
	\end{figure}
	
	\noindent \textbf{w/o voxel-based correction}
	After inferring skeleton from our first stage, it is a straightforward approach to acquire a base mesh by directly applying the corrosion technique for volume generation, and the base mesh can be extracted. The visual comparisons of this method against ours are shown in Fig \ref{fig:wo_voxel}. It can be seen, without volume correction, the wrong predictions caused by skeleton inference will be transferred to the mesh refinement stage, affecting the final output. Our proposed voxel-based correction network addresses this issue effectively.

	\subsection{Evaluation on Skeleton Inference}
	\label{SecCompSke}

	In this section, we conduct comparisons with several variants of our skeleton inference approach, to verify our final model is the optimal choice. These variants include: "Point-only fitting" method directly adopts PSG ~\cite{fan2017point} to regress the skeletal points; "Line-only fitting" method removes the square stream of our model and only deforms multiple lines to approximate the skeleton; "Square-only fitting" removes the line stream of our model and deforms multiple squares to fit the skeleton; "Line-and-Square fitting" method learns the deformation of multiple lines and squares together using a single MLP to approximate the skeleton; "Ours w/o laplacian" stands for our model without laplacian smoothness term. Note that, laplacian smoothness loss is also used for the training of "Line-only fitting", "Square-only fitting" and "Line-and-Square fitting".
	
	\begin{table}[h]
		\begin{center}
			\begin{tabular}{c | c}\hline
				\centering
				Methods & CD  \\
				\hline
				Point-only fitting & 1.185  \\
				Line-only fitting & 1.649  \\
				Square-only fitting & 1.185 \\
				Line-and-Square fitting & 1.252  \\
				Ours w/o laplacian & 1.621  \\
				Ours & \textbf{1.103}  \\
				\hline
			\end{tabular}
			\caption{The quantitative comparisons on the variants of our skeleton inference method. The Chamfer Distance($\times$ $10^3$) are reported.}
			\label{tab:ske_metric}
		\end{center}
	\end{table}
	
	\noindent \textbf{Quantitative results} All of these methods are evaluated on CD metric and the results are shown in Tab \ref{tab:ske_metric}. It can be seen that our final model outperforms all the others. Another discovery is that laplacian regularizer is very helpful to reach better accuracy.
	
	\noindent \textbf{Qualitative results} We then report the visual comparisons of these methods on a sampled example in Fig \ref{fig:com_ske}. As shown, point-only fitting results in scattered points no mention to the structures. Line only fitting fails to recover the surface-shaped skeleton parts. Square-only fitting can not capture the long and thin rods and legs. The method of Line-and-Square fitting causes messy outputs since a single MLP is difficult to approximate diverse local structures. As observed, the involvement of laplacian loss effectively improves the visual appearance of the results.
	
	\section{Conclusion}
	Recovering the 3D shape of an object from one of its perspectives is a very fundamental yet challenging task in computer vision field.
	The proposed framework splits this challenge task into three stages. It firstly recovers a 3D meso-skeleton represented as points, these skeletal points are then converted to its volumetric representation and passed to a 3DCNN for a solid volume synthesis. From which, a coarse mesh can be extracted. A GCNN is finally trained to learn the mesh deformation for producing geometric details. As demonstrated in our experiments both qualitatively and quantitatively, the proposed pipeline outperforms all existing methods. There are two directions worth being explored in the future: 1)how to change the whole pipeline to be an end-to-end network; 2) trying to apply adversarial learning on skeletal point inference, volume generation, and mesh refinement, for further improving the quality of final output mesh.
	
	\section{Acknowledge}
	This work is supported in part by the National Natural Science Foundation of China (Grant No.: 61771201), the Program for Guangdong Introducing Innovative and Enterpreneurial Teams (Grant No.: 2017ZT07X183), the Pearl River Talent Recruitment Program Innovative and Entrepreneurial Teams in 2017 (Grant No.: 2017ZT07X152), and the Shenzhen Fundamental Research Fund (Grants No.: KQTD2015033114415450 and ZDSYS201707251409055).
	
	{\small
		\bibliographystyle{ieee}
		\bibliography{egbib}
	}
	%
		\begin{appendix}
		\clearpage
		\noindent \large \textbf{Supplementary Material}\\
		\setlength{\tabcolsep}{4pt}
		\begin{table}[b]
			\begin{center}
				\begin{tabular}{c|*{5}{c}}\hline
					\centering
					Method & car& cab.& cou.& lam.& wat. \\
					\hline
					AtlasNet &1.320 &0.848 & 1.358 & 3.999 & 2.560  \\
					Pixel2Mesh & 1.408 & 0.893 & 1.376 & \textbf{3.561} & 1.867 \\
					Ours    & \textbf{0.717} & \textbf{0.708} & \textbf{1.350} & 3.639 &\textbf{1.597}\\
					\hline
					\hline
					
					Method & car & cab. & cou. & lam. & wat. \\
					\hline
					AtlasNet &8.035 &6.366 & 7.390 & 10.750 & 9.385 \\
					Pixel2Mesh &5.706 &4.307 & \textbf{5.490} & 10.176 & 6.671\\
					Ours     & \textbf{4.765} & \textbf{4.277} & 5.770 & \textbf{9.990} &\textbf{6.408} \\
					\hline
				\end{tabular}
				\vspace{0.01cm}
				\caption{Evaluations of our method on the other five shape categories. The top rows are the results measured by Chamfer distance($\times$ $10^3$) and the bottom rows are the results in Earth Mover's distance($\times$ 0.01). }
				\label{tab:cd_emd_sup}
			\end{center}
		\end{table}
		
		In this supplementary material, we conduct comparisons with state-of-art methods on more categories in Tab \ref{tab:cd_emd_sup} and provide additional metrics defined on meshes in Tab \ref{tab:metro}. We present more experimental results: Fig \ref{fig:more_recon} shows more comparisons of our method against state-of-the-arts. A set of input-output pairs are also shown in Fig \ref{fig:gallery}.
		
		\begin{table}[b]
			\begin{center}
				\begin{tabular}{c|*{5}{c}}\hline
					\centering
					Methods & pla. & ben. & cha. & tab. & fir.\\
					\hline
					AtlasNet  & 1.856  & 1.258 & 1.093 & 1.441 & 1.635\\
					Pixel2Mesh & \textbf{1.477} & 1.267 & 1.246 & 1.360 & 1.117 \\
					Ours & 1.643 &  \textbf{1.157} & \textbf{0.948} & \textbf{0.980} & \textbf{0.988} \\
					\hline
					\hline
					Methods  & car & cab. & cou. & lam. & wat. \\
					\hline
					AtlasNet & 1.528 & 1.105 & \textbf{1.145} & 1.865 & 1.633\\
					Pixel2Mesh & 1.230 & 1.113 & 1.209 & \textbf{1.172} & 1.315\\
					Ours     & \textbf{1.190}  &\textbf{1.028} & 1.270 & \textbf{1.172}  & \textbf{1.245}\\
					\hline
				\end{tabular}
				\vspace{0.15cm}
				\caption{The results measured by Metro distance($\times$ $10$). The lower the results are better.}
				\label{tab:metro}
			\end{center}
		\end{table}
		\setlength{\tabcolsep}{1.4pt}
		
		\section{More evaluations on other categories}
		\label{SecMoreCat}
		
		To demonstrate the superiority of method against the state-of-art methods, we conduct more quantitative comparisons by CD and EMD on other five categories. We select other five categories in ShapeNet \cite{chang2015shapenet}: car(4000), cabinet(1572), couch(3172), lamp(1956), watercraft(1658). The training and testing paradigms are consistent with details described in the previous text.  For simplicity, only the results of  AtlasNet \cite{groueix2018atlasnet} and Pixel2Mesh \cite{wang2018pixel2mesh} are reported.
		
		\section{Comparisons on metro distance}
		\label{SecComMet}
		
		Metro distance \cite{cignoni1998metro} is defined as the hausdorff distance between point clouds sampled from the true and generated meshes. Tab \ref{tab:metro} lists the quantitative comparisons measured by Metro distance (we exactly follow the method presented in AtlasNet \cite{groueix2018atlasnet} for the evaluation) on all 10 categories. Note that in both evaluations, our method outperforms the others in almost all categories.
		
		\section{More Qualitative comparisons}
		\label{SecMoreQua}
\begin{figure*}[t]
			\begin{center}
				\includegraphics[height=0.92\textheight]{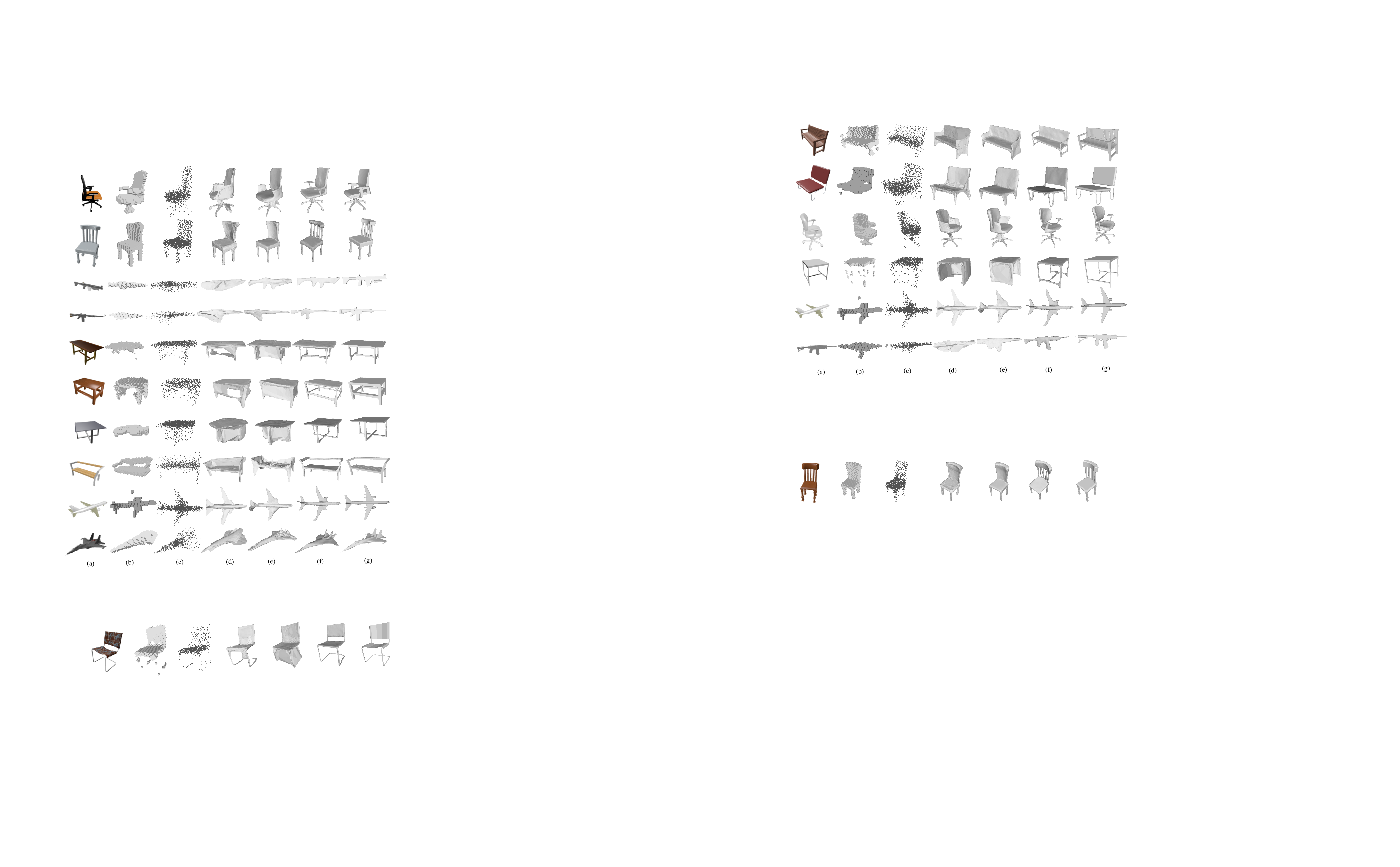}\\
			\end{center}
			\caption{(a)Input images; (b)R2N2;  (c)PSG;  (d)AtlasNet; (e)Pixel2Mesh;  (f)Ours;  (g)Ground truth}
			\label{fig:more_recon}
		\end{figure*}

		In this section, we show more qualitative comparison results against state-of-the-arts. As shown in Fig \ref{fig:more_recon}, for objects with more complex topology (i.e. non-zero genus), our method can better reconstruct the holes and loops of these 3D objects than other state-of-arts.
		
		\begin{figure*}[t]
			\begin{center}
				\includegraphics[width=0.95\linewidth]{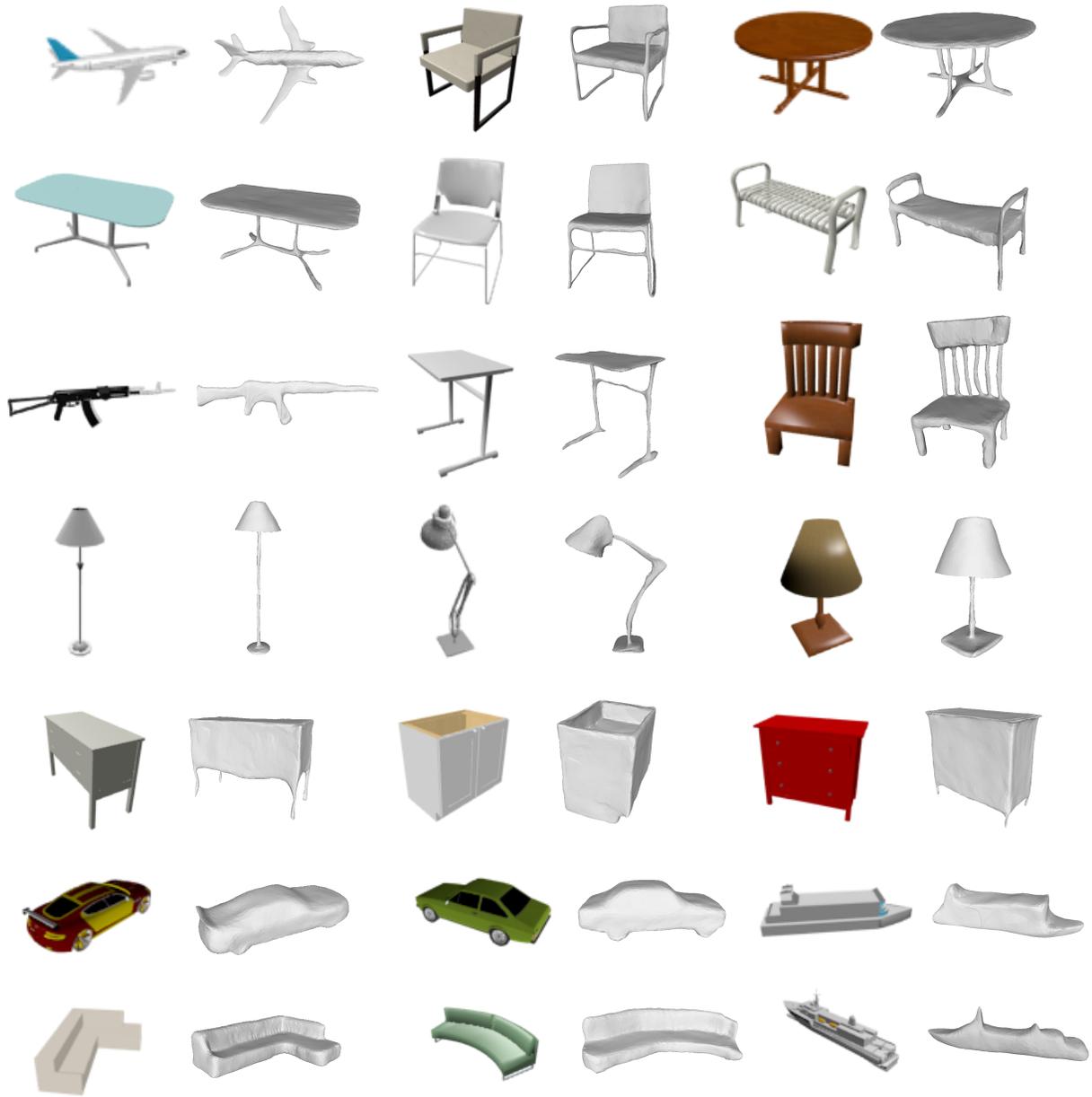}\\
			\end{center}
			\caption{More input-output pairs of our approach are shown. We also show our results on some other categories like: car, cabinet, couch, lamp, watercraft. }
			\label{fig:gallery}
		\end{figure*}

		\section{Results gallery}
		\label{SecGal}
		A set of input-output pairs are also shown in Fig \ref{fig:gallery}.
			
		\end{appendix}
\end{document}